\documentclass[conference]{IEEEtran}
\IEEEoverridecommandlockouts
\usepackage{cite}
\usepackage{amsmath,amssymb,amsfonts}
\usepackage{graphicx}
\usepackage{textcomp}
\usepackage{xcolor}

\usepackage{algorithm}
\usepackage{algorithmicx}  
\usepackage{algpseudocode}  
\usepackage{amsmath}

\def\BibTeX{{\rm B\kern-.05em{\sc i\kern-.025em b}\kern-.08em
    T\kern-.1667em\lower.7ex\hbox{E}\kern-.125emX}}
\begin{document}

\title{Efficient Urban-scale Point Clouds Segmentation with BEV Projection

\author{
\IEEEauthorblockN{Zhenhong Zou}
\IEEEauthorblockA{\textit{School of Vehicle and Mobility} \\
\textit{Tsinghua University}\\
Beijing, China \\
zouzhehong@mail.tsinghua.edu.cn}
\and
\IEEEauthorblockN{Yizhe Li}
\IEEEauthorblockA{\textit{College of Information Science and Engineering} \\
\textit{Xinjiang University}\\
Urumqi, China \\
liyizhe2001@126.com}
}
}


\maketitle

\begin{abstract}
Point clouds analysis has grasped researchers' eyes in recent years, while 3D semantic segmentation remains a problem. Most deep point clouds models directly conduct learning on 3D point clouds, which will suffer from the severe sparsity and extreme data processing load in urban-scale data. To tackle the challenge, we propose to transfer the 3D point clouds to dense bird's-eye-view projection. In this case, the segmentation task is simplified because of class unbalance reduction and the feasibility of leveraging various 2D segmentation methods. We further design an attention-based fusion network that can conduct multi-modal learning on the projected images. Finally, the 2D out are remapped to generate 3D semantic segmentation results. To demonstrate the benefits of our method, we conduct various experiments on the SensatUrban dataset, in which our model presents competitive evaluation results (61.17\% mIoU and 91.37\% OverallAccuracy). We hope our work can inspire further exploration in point cloud analysis.

\end{abstract}



\section{Introduction}
3D semantic segmentation is the critical technology of point cloud learning with the purpose of assigning a semantic label to each individual point data, which has been extensively applied in automatic driving \cite{Cortinhal2020SalsaNextFS}, virtual reality \cite{Hu2020RandLANetES}, 3D reconstruction \cite{Hu2020TowardsSS}, etc. Although deep learning has prominent performance in 2D semantic segmentation tasks, it unable to directly process the point data which is irregular, unordered and unstructured \cite{Xu2020SqueezeSegV3SC}.
Therefore, several methods \cite{Su2018SPLATNetSL, Rosu2020LatticeNetFP, Graham20183DSS,Tchapmi2017SEGCloudSS,Tatarchenko2018TangentCF,Lawin2017DeepP3,Wu2019SqueezeSegV2IM} currently convert unstructured points into certain efficient intermediate representations, such as voxels \cite{Huang2016PointCL,Graham20183DSS}and multi-views\cite{Lawin2017DeepP3,Tosteberg2017SemanticSO,Boulch2017UnstructuredPC}, to process point clouds utilizing classic CNN models.
With the increasing demand for 3D scene understanding, more and more 3D point cloud datasets are proposed. From the indoor datasets (e.g., S3DIS \cite{Armeni2017Joint2D} and ScanNet \cite{Dai2017ScanNetR3}) to roadway-level datasets (e,g., SemanticKITTI \cite{Behley2019SemanticKITTIAD}), the spatial size of datasets is also larger. Recent work \cite{Hu2020TowardsSS,Li2020Campus3DAP} have proposed urban-level datasets which bringing several fire-new challenges to semantic segmentation of large-scale datasets. 

\begin{figure}[t!]
\centering
\includegraphics[width=0.5\textwidth]{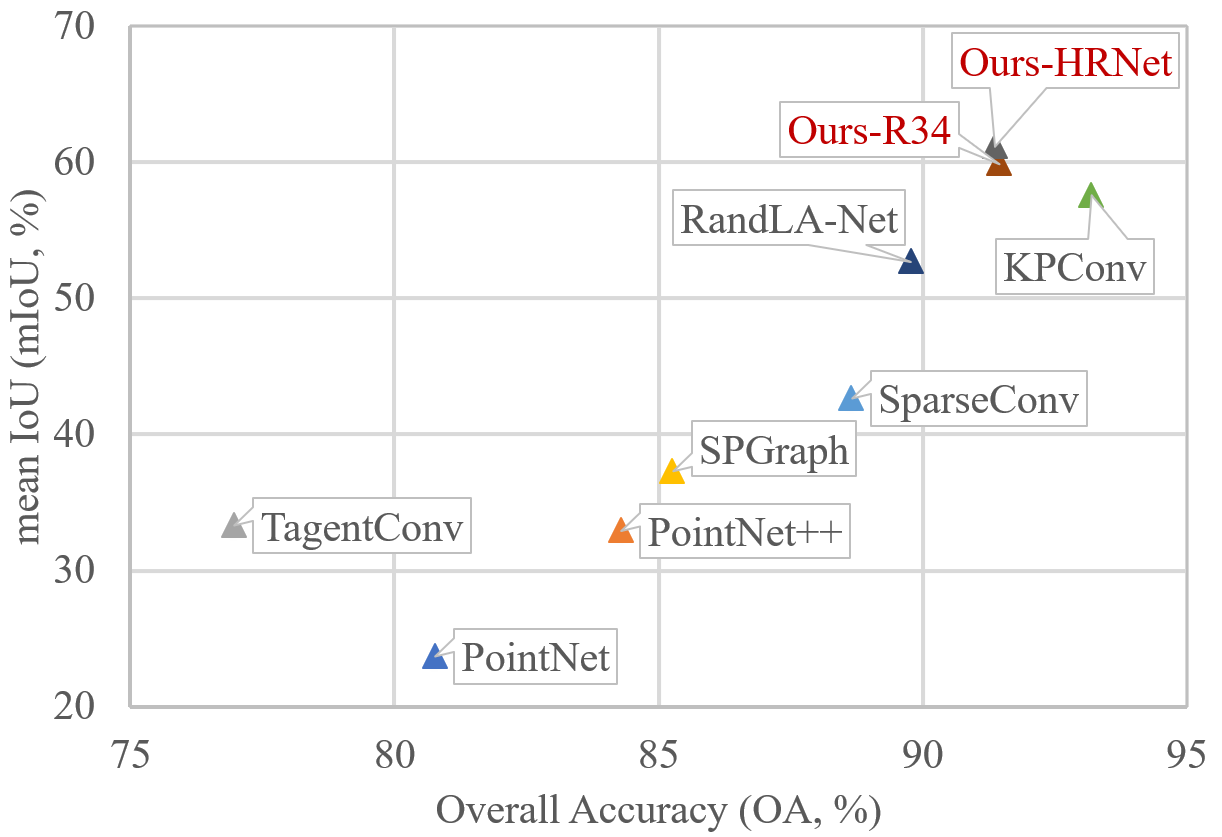}
\caption{Comparison results with other published methods. We implemented our models using ResNet-34 and HRNet, both of which achieved competitive mean IoU and overall accuracy on the SensatUrban dataset. Notice that our results are computed on the validation set due to the unaccessibility of test set labels, while other models' results are provided by the dataset publishers and evaluated with testing data.}
\label{compare}
\end{figure}

Different from the LiDAR-based datasets, these urban-scale point cloud are mostly obtained from UAV photogrammetry, which may lead to the the following characteristics in the dataset. Firstly, the scanning of UAV Photogrammetry is uneven, the scanning area is not concentrated, and the captured images include scattered areas on the edges. Secondly, the reconstructed point cloud is partially missing. We observed this phenomenon in the SensatUrban \cite{Hu2020TowardsSS} dataset, a typical example is that, after visualizing the point cloud, there is no corresponding wall points under the roof, which makes the roof seem to be suspended in the air.
Interestingly, We found that the category overlap rate of the vertical points in UAV-based point clouds is lower, e.g., SensatUrban is 2.3$\%$, meaning that bird's-eye-view is a suitable projection method which is simpler, more efficient, and is able to  maximize retain the point details. Moreover, for the projected images, the 2D pixel-level dataset with richer markers can be used for pre-training. Therefore, in this paper, we propose a BEV projection segmentation methods to deal with urban-scale 3D segmentation problem. Our main contributions are: 1) conduct point-level analysis for urban-scale point clouds; 2) propose a multi-modal fusion segmentation model with specific BEV projection algorithm; 3) we evaluate our method in the SensatUrban dataset, our competitive result proves the efficiency of our design.

\begin{figure}[t!]
\centering
\includegraphics[width=0.5\textwidth]{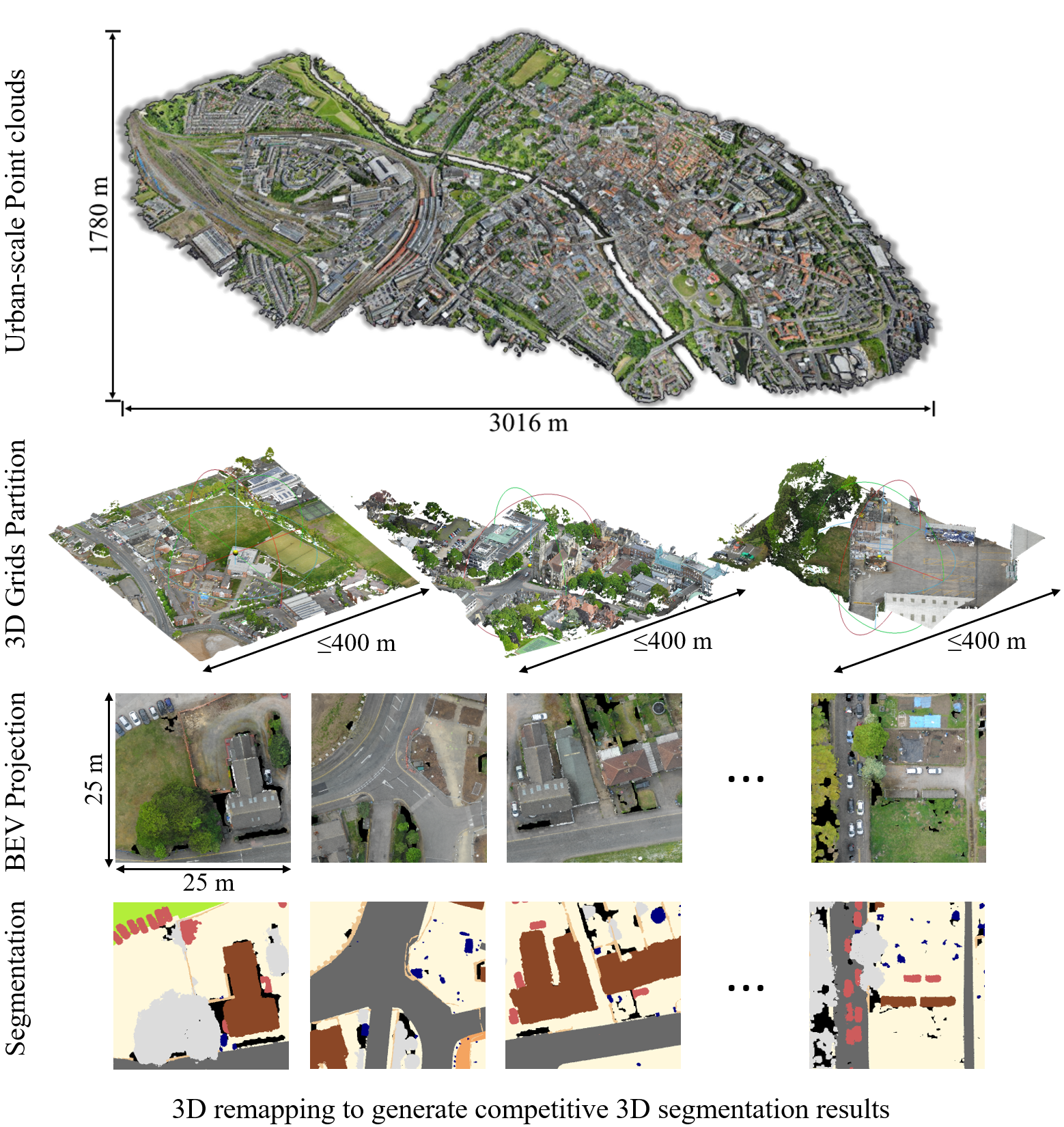}
\caption{Overview of the 3D-2D task transfer process. The urban-scale point clouds map are previously partitioned into grids with side length under 400 meters. We further generated squares of $25\times25 m^2$ with $20x$ magnification. The 2D segmentation output will be remapped to the large 3D map according to their x/y coordinates.}
\label{overview}
\end{figure}

\section{Related Works}
\subsection{3D Semantic Segmentation}
In general, according to the form of point cloud data entered into the network, most of the existing 3D semantic segmentation approaches can be divided into three categories: point-based, 3D representation-based and projection-based. 

Point-based approaches directly deal with the raw point cloud, and the representative methods of it is PointNet which has high computation overhead. 
Although \cite{Qi2017PointNetDL,Li2018PointCNNCO} have made some beneficial improvements to PointNet, it is still hard to accelerate because these methods deal with the sprase data directly.
Recent RandLA-Net \cite{Hu2020RandLANetES} introduces random sampling and a light network architecture which speed the model greatly. However, as mentioned in \cite{Liu2019PointVoxelCF}, an issue couldn't be ignored in point-based method is that the large time waste (80$\%$) of processing sprase data caused by inefficient random memory access, meaning only small amount of time is actually used to extract features. In addition, large memory overhead is also a severe problem.

3D representation-based approaches transform the raw point cloud data into certain 3D representation (e.g., voxels and lattices) and then leverage 3D convolution  
\cite{Su2018SPLATNetSL, Rosu2020LatticeNetFP, Graham20183DSS,Tchapmi2017SEGCloudSS}. 
However, it is difficult to balance the relationship between resolution and memory \cite{Guo2020DeepLF,Liu2019PointVoxelCF}. The lower resolution, the more serious information loss of point cloud because the points in the same grid are merged together. The higher resolution, the greater computational overhead and memory usage. Moreover, the pre-processing and post-processing steps require a lot of time \cite{Qiu2021SemanticSF}. 

Instead processing points directly, projection-based approach utilizes mature 2D convolution models to process images projected from 3D point cloud.
Projection-based approaches include several specific categories such as multi-view, spherical-based method.
Multi-view method \cite{Lawin2017DeepP3,Tosteberg2017SemanticSO,Boulch2017UnstructuredPC} projects point clouds into multiply virtual camera view. 
For example, \cite{Lawin2017DeepP3} utilizes a multi-stream CNN to process images generated from each view and then fuse the prediction scores from different images of each point, \cite{Tosteberg2017SemanticSO} defines a rotated camera and proposes Katz projection to choose points in each camera angle, \cite{Boulch2017UnstructuredPC} generates depth images and RGB images in different camera positions. 
\cite{Wu2018SqueezeSegCN} utilize spherical projection method to transform the 3D point cloud into the images, uses the SqueezeSeg network for segmentation and applies CRF (Conditional Random Field) to optimize the segmentation results. \cite{Wu2019SqueezeSegV2IM} proposes Context Aggregation Module (CAM) to expand receptive field based on SqueezeSeg and \cite{Xu2020SqueezeSegV3SC} introduces Spatially-Adaptive Convolution (SAC) to further improve the segmentation accuracy. 

\subsection{Semantic Segmentation for Large-scale Scene }
Several urban-scale 3D point cloud datasets 
\cite{Hu2020TowardsSS,Li2020Campus3DAP,Can2020SemanticSO}
photographed by UAV
has been proposed in recent work 
and the largest among them is the SensatUrban \cite{Hu2020TowardsSS} dataset, which covers the area of 7.64×10$^{6}$ m$^{2}$, with 3 billion annotated points. However, these large and dense datasets bring new challenges to semantic segmentation.

Firstly, facing the massive data, the choice of pre-process methods, e.g., data partition, down-sampling, etc. is of great significance.
Secondly, urban-scale point cloud exists the problem of the imbalance class distribution. 
Thirdly, a significant difference between the UAV-based datasets and Lidar-based datasets is that the former contains the RGB features. 
For large-scale datasets, whether to incorporate RGB features into the network and how to utilize RGB features efficiently is worth considering. Recent work, e.g., RandLA-Net \cite{Hu2020RandLANetES} and BAAF-Net \cite{Qiu2021SemanticSF} have utilized the RGB color and achieved positive segmentation results.
For the images generated by BEV projection, we design a multi-model fusion network based on attention, which fuses RGB and gometric details efficiently. Compared with the single-modal network, the segmentation effect achieves a certain improvement, which further verifies the significance of RGB color for segmentation.

Rencently, several semantic segmentation algorithms for large dataset have been proposed \cite{Hu2020RandLANetES,Liu2020FGNetFL,Tatarchenko2018TangentCF,Landrieu2018LargeScalePC}.
For example, RandLA-Net \cite{Hu2020RandLANetES} introduces the random sampling to improve efficiency of computation and memory, TagentConv \cite{Tatarchenko2018TangentCF} utilizes a U-type network based on tangent convolution for the semantic segmentation of large and dense datasets and SPGraph \cite{Landrieu2018LargeScalePC} proposes a superpoint graph (SPG), a novel point cloud representation which is capable to capture the contextual structure of 3D points. More large-scale point cloud segmentation algorithms need to be proposed.

\section{Proposed Method}

\subsection{Problem Statement}
The purpose of 3D point cloud semantic segmentation is to assign a semantic label to each individual point, while 2D segmentation is to assign a specific label to each pixel. To some extent, these two types of tasks have similar purposes and solutions.
According to our statement above, the 3D point clouds semantic segmentation task can be transferred to a 2D Bird's-eye-view Segmentation problem. The main processes include Bird's-eye-view Mapping and 2D multi-modal segmentation.

\begin{algorithm}[b!]  
  \caption{Bird's-eye-view Projection \& Completion}  
  \label{BEVprojection}  
  \begin{algorithmic}[1]  
    \Require  
      $X$: point clouds set;
      $f(x)$: point completion function;   \newline
      $g_{scale}$: magnified scale;
      $g_{size}$: grid size;
      $g_{step}$: step size; 
    \Ensure  
      $B$: BEV projected multi-modal images $RGB$, $Alt$
    \For{$X_i \in X$}
        \State $x_{min}, y_{min}\gets X_i$;
        \For{$(x_i,y_i,z_i) \in X_i$}
            \State $x_i\gets int((x_i-x_{min})/g_{scale})$;
            \State $y_i\gets int((y_i-y_{min})/g_{scale})$;
        \EndFor
        \State $x_{max}, y_{max}\gets X_i$;
        \For{$x_s \geq 0; x_s \leq x_{max}; x_s++; g_{step}$}
            \For{$y_s \geq 0; y_s \leq y_{max}; y_s++; g_{step}$}
                \State initialize grids;
                \If{$x_i \leq x_s+g_{size}$}
                    \State grids$\gets(x_i,y_i,z_i)$;
                \EndIf
                \For{$(x_i,y_i,z_i) \in grids$}
                    \State select top $z_i$ for $x_i,y_i$;
                    \State remove other points;
                    \State $RGB[x_i,y_i]\gets X_i[x_i,y_i,z_i]$;
                    \State $Alt[x_i,y_i]\gets z_i$;
                \EndFor
                \State initialize $RGB$, $Alt$;  
                \State $RGB\gets$ $f$($RGB$, (3,3));
                \State $Alt\gets$ $f$($Alt$, (3,3));
                \State return $RGB, Alt, x_s, y_s$;
            \EndFor
        \EndFor
    \EndFor
  \end{algorithmic}  
\end{algorithm}

\subsection{Why Bird's-eye-view projection is reasonable}
It requires the consistency in input data and expected output when we transfer one task to another. To evaluate our idea, we conduct point level analysis before building models. We first project the 3D points onto the BEV map (which will be detailed in the following) and count the overlap ratio. When we scale the coordinates at 0.04m in projection, about 25.44\% points will be lost. For those places with dense points, the ratio will raise up to 50\% or more. However, we found that most of the overlapped points belongs to the same categories as the top points. The class overlapped ratio is lower than 2.3\% and mIoU can be up to 93.7\%. In this case, it is possible to transfer the 3D segmentation task to 2D BEV segmentation. And our goal will be precise recognition on the BEV images.

\subsection{Bird's-eye-view Mapping}
\begin{figure}[t!]
\centering
\includegraphics[width=0.5\textwidth]{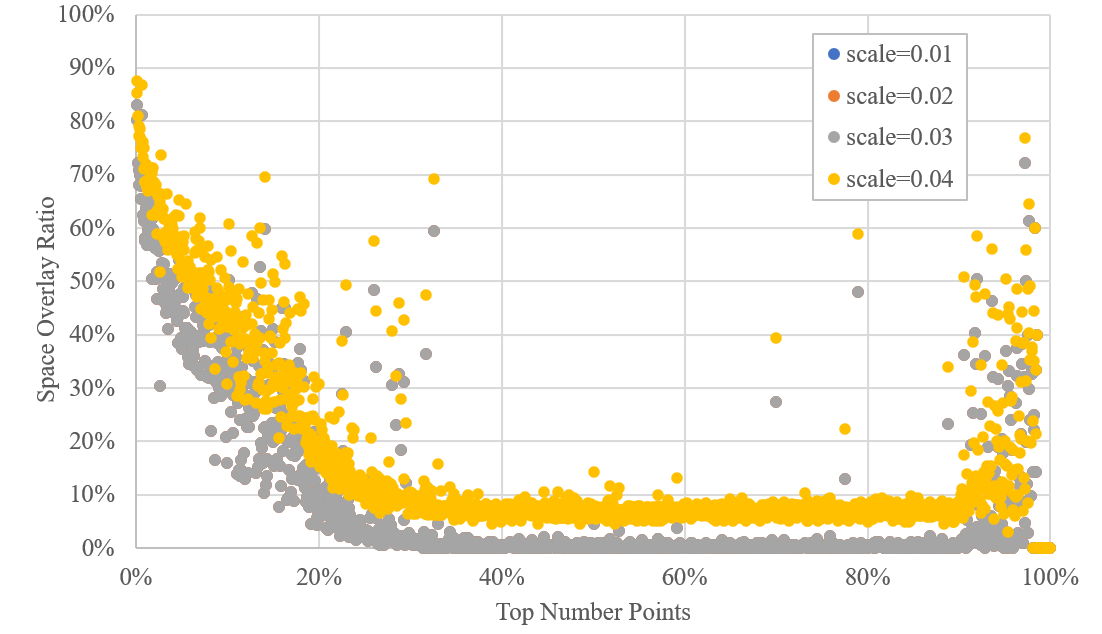}
\caption{Spatial overlap ratio statistical result. The x-axis indicates the rank of sliding windows ($1m\times 1m$) by their point numbers, top 0\% means windows with the most points. The y-axis indicates the overlap ratio of points projected onto BEV images.}
\label{overlap}
\end{figure}

In order to optimize the data processing load of such large point clouds, we separate the whole work into three parts: 3D-to-BEV projection, sparse BEV images completion, and BEV-to-3D remapping. The processes of the first two parts are presented as pseudo code in the Algorithm \ref{BEVprojection} below. We set a sliding window to process points and generate BEV images. Before projection, we need to initialize the parameters $g_{scale}$, $g_{scale}$, $g_{scale}$, that controls the scale, size, and moving steps of the sliding windows. For each sliding step, we sort the points by x/y coordinates, and query points from current BEV projection window start/end coordinates, after which processed points will be removed to reduce the following data processing volume. To obtain the optimal parameters, we test the spatial overlap ratio with different projected scales from 0.01 to 0.04, as shown in the Fig.\ref{overlap}. When we set the scale in [0.01,0.03], it will result in the very close overlap distribution for different parts of the point clouds, that means the minimum interval of points in urban-scale point clouds is within [0.03,0.05](m). Besides, the proper length of windows is within [20,50](m) according to our projected images number estimation. Therefore, we set the parameters as $g_{scale}=0.05$, $g_{size}=g_{step}=25$. However, we also suggest multi-scale, multi-size, and multi-step sampling for better training in the future work or other similar tasks.

\begin{figure}[htb!]
\centering
\includegraphics[width=0.48\textwidth]{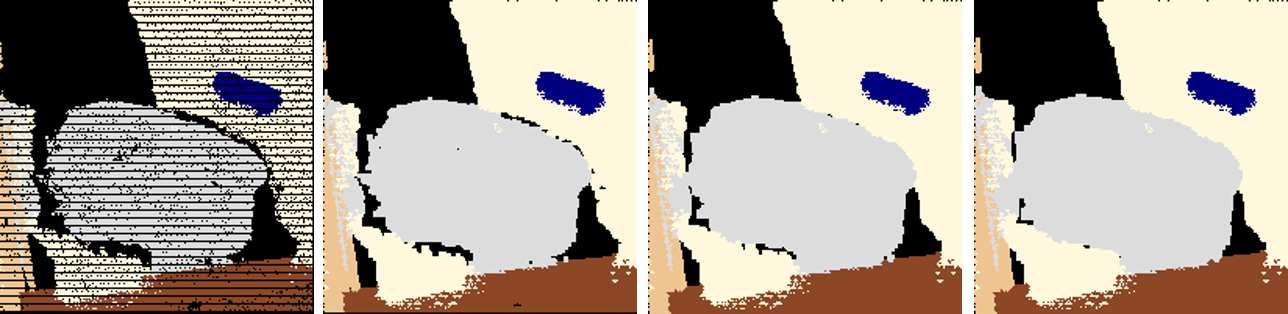}
\caption{Completion test. For left to right, we present the raw BEV label and label with one/two/three times maxpooling completion.}
\label{completion}
\end{figure}

As for points in a single sliding window, we map the points to the pixels by integralizing the x/y coordinates. Inevitably it will bring the loss of value quantization, however, it will not affect the label retrieve process if we conduct the same process in 3D remapping. The BEV map is updated with points at the top, generating the RGB and Altitude(Alt) images with their color and z-coordinate values. Considering the significant sparsity of projected point clouds on the BEV images, which will introduce severe noise into the label and model learning, it is necessary to conduct pixel-level completion for the projection, especially for the internal area and margin around different class points. In our experiments, we iteratively conduct three times 2D maxpooling for each channels in each images. The progressive changes in labels are shown in Fig.\ref{completion}.

As for 3D remapping, we store the absolute x/y coordinate for every projected windows, and use theme to query the extract places in the raw large-scale point clouds for the 2D segmentation outputs. Points corresponding to the same pixel will be valued the same classes as the pixels. After that, we are able to evaluate the 3D semantic segmentation performance.

\subsection{Multi-modal Segmentation}
With the altitude and RGB images from the BEV projection, we can leverage a multi-modal network to learn from different aspects of the data. In order to quickly develop a suitable model, we consider an Encoder-Decoder network UNet as our baseline not only for its popular model architecture, but also for its efficiency in modification, training, and inference. It comprises 4 blocks in the encoder and 5 in the decoder, in which two are ResNet-34 blocks, the last four layers use transposed convolution and the rest are convolutional blocks. All convolutional blocks have a batch-normalization layer and a ReLU layer following the convolution layer, and all kernels size are 3x3. Each block in the encoder is linked to the corresponding blocks in the decoder with a dash line, that concatenate the output of them to retrieve low-level features.

Generally, multi-modal fusion depends on the feature communication in various layers. Driven by it, we proposed a flexible multi-stage fusion network, which supports different times and places to fusion multi-pipeline data. The fusion layers comprise several constant shape fusion blocks. Each block accepts two equal-shape tensors from two pipelines, and adopts an attention layer to select the key channel from the concatenated feature map. In this way, the fusion blocks tend to drop the unrelated features and fuse those are easy to be activated in the following layers. For the attention block, we refer to our previous work\ref{Zhang2021ChannelAI}, which proposed a cross-channel multi-modal fusion attention block for semantic segmentation. After that, we add a 1x1 convolution to reduce the dimensionality, and repeat such a fusion block for image feature and fusion features, altitude features and fusion features. It is important that out block keeps the constant shape of feature maps, that mean we can stacked unlimited blocks with various network shape as we want.


\begin{table*}[htp!]
\centering
\addtolength{\leftskip} {-2cm}
\addtolength{\rightskip}{-2cm}
\renewcommand\arraystretch{1.2}
\caption{Detailed comparison results on the SensatUrban dataset. * marks the results on validation set. Others are on the test set and provied by the data publishers.}
\begin{tabular}{lccccccccccccccccc}
\hline
 & 
\multicolumn{1}{c}{\rotatebox{60}{\textbf{OA(\%)}}} & 
\multicolumn{1}{c}{\rotatebox{60}{\textbf{mAcc(\%) }}} & 
\multicolumn{1}{c}{\rotatebox{60}{\textbf{mIoU(\%) }}} & 
\multicolumn{1}{c}{\rotatebox{60}{\textbf{ground}}} & 
\multicolumn{1}{c}{\rotatebox{60}{\textbf{veg.}}} & 
\multicolumn{1}{c}{\rotatebox{60}{\textbf{building}}} & 
\multicolumn{1}{c}{\rotatebox{60}{\textbf{wall}}} & 
\multicolumn{1}{c}{\rotatebox{60}{\textbf{bridge}}} & 
\multicolumn{1}{c}{\rotatebox{60}{\textbf{parking}}} & 
\multicolumn{1}{c}{\rotatebox{60}{\textbf{rail}}} & 
\multicolumn{1}{c}{\rotatebox{60}{\textbf{traffic.}}} & 
\multicolumn{1}{c}{\rotatebox{60}{\textbf{street.}}} & 
\multicolumn{1}{c}{\rotatebox{60}{\textbf{car}}} & 
\multicolumn{1}{c}{\rotatebox{60}{\textbf{footpath}}} & 
\multicolumn{1}{c}{\rotatebox{60}{\textbf{bike}}} & 
\multicolumn{1}{c}{\rotatebox{60}{\textbf{water}}} \\ \hline
PointNet\cite{Qi2017PointNetDL}       & 80.78  & 30.32    & 23.71    & 67.96  & 89.52  & 80.05    & 0.00   & 0.00   & 3.95    & 0.00   & 31.55    & 0.00    & 35.14  & 0.00     & 0.00   & 0.00  \\ 
PointNet++\cite{Qi2017PointNetDH}     & 84.30  & 39.97    & 32.92    & 72.46  & 94.24  & 84.77    & 2.72   & 2.09   & 25.79   & 0.00   & 31.54    & 11.42   & 38.84  & 7.12     & 0.00   & 56.93   \\ 
TagentConv\cite{Tatarchenko2018TangentCF}     & 76.97  & 43.71    & 33.30    & 71.54  & 91.38  & 75.90    & 35.22  & 0.00   & 45.34   & 0.00   & 26.69    & 19.24   & 67.58  & 0.01     & 0.00   & 0.00  \\
SPGraph\cite{Landrieu2018LargeScalePC}        & 85.27  & 44.39    & 37.29    & 69.93  & 94.55  & 88.87    & 32.83  & 12.58  & 15.77   & 15.48  & 30.63    & 22.96   & 56.42  & 0.54     & 0.00   & 44.24   \\
SparseConv\cite{Graham20183DSS}     & 88.66  & 63.28    & 42.66    & 74.10  & 97.90  & 94.20    & 63.30  & 7.50   & 24.20   & 0.00   & 30.10    & 34.00   & 74.40  & 0.00     & 0.00   & 54.80   \\
KPConv\cite{Thomas2019KPConvFA}         & \textbf{93.20}  & 63.76    & 57.58    & 87.10  & 98.91  & 95.33    & 74.40  & 28.69  & 41.38   & 0.00   & 55.99    & 54.43   & 85.67  & 40.39    & 0.00   & 86.30   \\
RandLA-Net\cite{Hu2020RandLANetES}         & 89.78  & 69.64    & 52.69    & 80.11  & 98.07  & 91.58    & 48.88  & 40.75  & 51.62   & 0.00   & 56.67    & 33.23   & 80.14  & 32.63    & 0.00   & 71.31   \\ \hline
Ours-UNet-R34*  &  \underline{91.45} & 67.76 & 59.90 & 78.45 & 95.39 & 97.89 & 60.72 & 59.80 & 40.45 & 40.35 & 67.32 & 51.42 & 82.11 & 41.94 & 0.00 & 62.87 \\ 
Ours-Deeplabv3-R101* & 90.53 & \multicolumn{1}{c}{\textbf{72.32}} & \underline{60.28} & \multicolumn{1}{c}{81.27} & \multicolumn{1}{c}{90.09} & \multicolumn{1}{c}{93.98} & \multicolumn{1}{c}{52.28} & \multicolumn{1}{c}{59.82} & \multicolumn{1}{c}{49.32} & \multicolumn{1}{c}{15.88} & \multicolumn{1}{c}{72.81} & \multicolumn{1}{c}{48.72} & \multicolumn{1}{c}{76.86} & \multicolumn{1}{c}{46.23} & \multicolumn{1}{c}{0.00} & \multicolumn{1}{c}{61.51} \\ 
Ours-OCRNet-HRNet*       & 91.37      & \multicolumn{1}{c}{\underline{71.87}} & \textbf{61.17} & \multicolumn{1}{c}{83.21} & \multicolumn{1}{c}{92.16} & \multicolumn{1}{c}{94.40} & \multicolumn{1}{c}{54.84} & \multicolumn{1}{c}{28.61} & \multicolumn{1}{c}{52.25} & \multicolumn{1}{c}{36.55} & \multicolumn{1}{c}{74.46} & \multicolumn{1}{c}{50.91} & \multicolumn{1}{c}{80.10} & \multicolumn{1}{c}{48.19} & \multicolumn{1}{c}{0.00} & \multicolumn{1}{c}{65.37} \\ \hline
\end{tabular}
\label{Result}
\end{table*}

\section{Experiments}

\subsection{Setup}
\subsubsection{Dataset}
SensatUrban \cite{Hu2020TowardsSS} captured in 3 large cities in UK  includes 2847M points and covers an area of 7.64×10$^{6}$ m$^{2}$ in real world, which is the largest 3D point cloud dataset at present. 
After obtaining areial image sequences captured by UAV, the SensatUrban point cloud dataset is reconstructed from these images. It contains 13 semantic classes, including major categories such as ground, building, traffic road and also several minor categories such as bike, rail and bridge. 
In the experiment, 37 point clouds are used for training and 6 point clouds are used for testing. Each point contains the features of 3D coordinates, RGB color and semantic class. Notice that due to the lack of test set labels, we randomly split the train set by 4:1, using 80\% data for training and 20\% for testing. All test data are not used in training.
\subsubsection{Metrics}
We compared our model with several bechmarks \cite{Qi2017PointNetDL, Qi2017PointNetDH,Tatarchenko2018TangentCF,Landrieu2018LargeScalePC,Graham20183DSS,Thomas2019KPConvFA,Hu2020RandLANetES} which utilize different approaches (e.g., point-based method, projection-based method, etc.) and are published recently. Mean IoU (mIoU) and Overall Accuracy (OA) are chose as evaluation indicators.
\subsubsection{Implementation}
We use CrossEntropy as the loss function in our training. Considering the imbalance in different classes, we use log inverse weights to adjust the loss in learning. We set the batch size as 8 and the input size as the projection size at 500x500. Our models are trained on two GPUs, RTX 3090 with 24G RAM and E5-2678v3 CPU. Besides, we use the following software setup: Ubuntu 16.04 64 bit Operating System, Python 3.6, gcc5.4.0, PyTorch 1.7 with CUDA 11.0 hardware acceleration.

\subsection{Results}
\begin{figure*}[t!]
\centering
\includegraphics[width=1\textwidth]{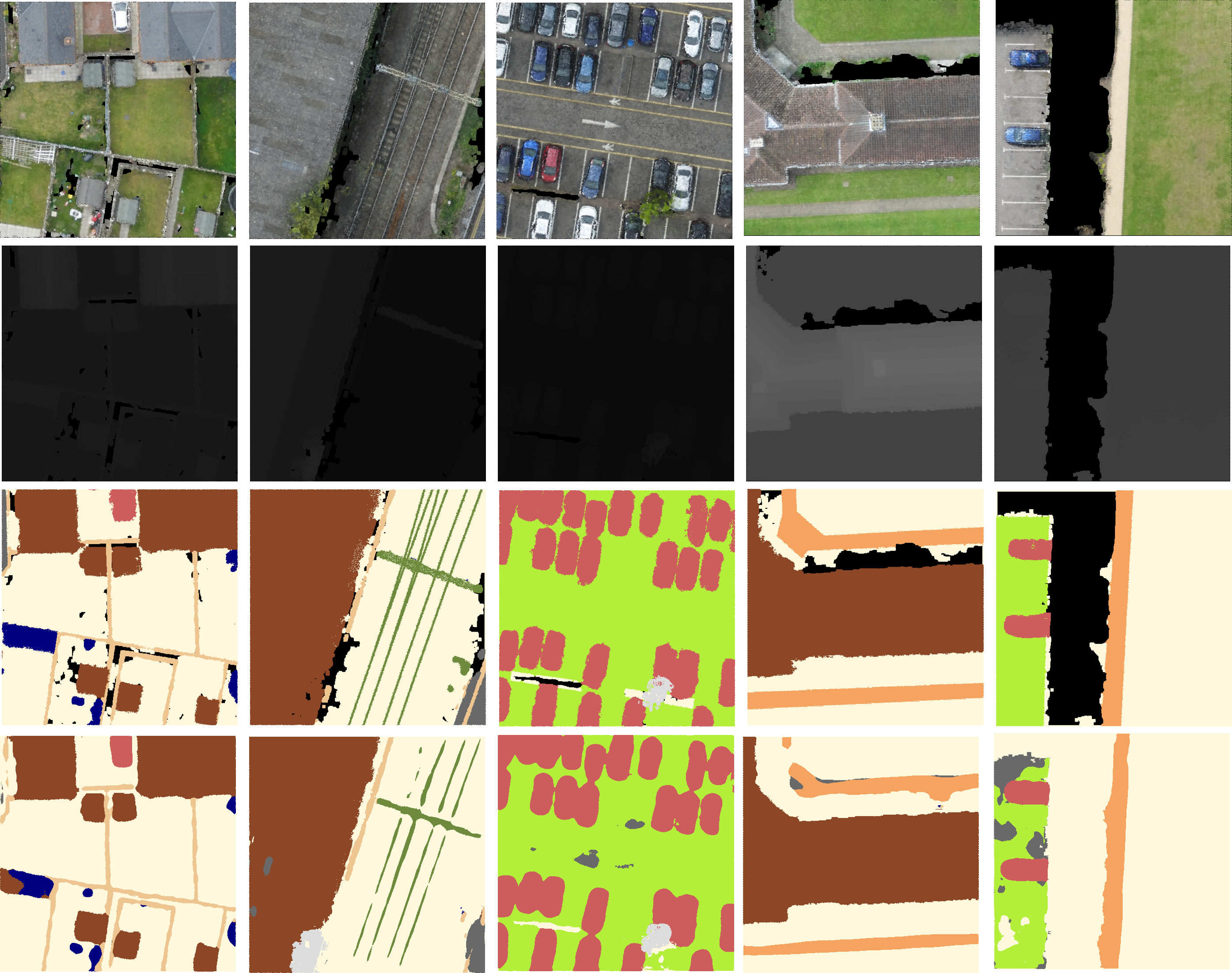}
\caption{Visualization of BEV segmentation. From top to down are: RGB images, altitude images, labels(dark  means invalid area without points), and 2D segmentation outputs.}
\label{visual}
\end{figure*}

We implementa our model using three backbones, UNet and ResNet34, Deeplabv3 and ResNet101, OCRNet and HRNet. The last two models are trained to explore the potential performance under our BEV segmentation framework. We have presented our segmentation results(remapped to 3D point clouds and evaluate in 3D) in the Table.\ref{Result}. Comparing with existing models, our model can achieve rather competitive results in most classed and the overall performancee in OA, mAcc, and mIoU. The drawback is that our BEV segmentation still fail to recognize some of the small objects like bike, because they also occupy very limited pixels in the projection images. The problem may be solved by fuse 3D and our BEV models in the future work. The visualization are shown in the Fig.\ref{visual}.

\section{Conclusion}
In this paper, we designed a preprocessing method for large-scale UAV point clouds, that is, projecting 3D point clouds to dense bird's-eye views to solve the problems of data sparseness and serious data processing burden in large-scale datasets. In addition, we also propose an multi-modal fusion network based on attention to segment the generated 2D images, making full use of RGB color and geometric information. We have obtained 61.17\% mIoU and 91.37\% OverallAccuracy test results on the SensatUrban dataset. We hope that our work can inspire large-scale point cloud semantic segmentation task.

\bibliographystyle{unsrt}
\bibliography{ref}

\vspace{12pt}

\end{document}